\documentclass{article}

\usepackage{arxiv}

\usepackage[utf8]{inputenc} 
\usepackage[T1]{fontenc}    
\usepackage{hyperref}       
\usepackage{url}            
\usepackage{booktabs}       
\usepackage{amsfonts}       
\usepackage{nicefrac}       
\usepackage{microtype}      
\usepackage{lipsum}		

\usepackage{graphicx}
\usepackage{subcaption}
\usepackage{algorithm,amsmath,algpseudocode}

\title{Computational neuron-level analysis of human cortex cytoarchitecture}

\date{} 					

\author{
  Andrija \v{S}tajduhar\thanks{Corresponding author - University of Zagreb, School of Medicine, Croatian Institute for Brain Research, \v{S}alata 12, Zagreb 10000, Croatia, \texttt{andrija.stajduhar@hiim.hr}} \\
  Croatian Institute for Brain Research\\
  University of Zagreb, School of Medicine\\
  \texttt{andrija.stajduhar@hiim.hr} \\
   \And
 Tomislav Lipi\'{c} \\
 Laboratory for Machine Learning and Knowledge Representation \\
  Ru\dj er Bo\v{s}kovi\'{c} Institute\\
  \texttt{tlipic@irb.hr} \\
  \And
  Goran Sedmak \\
  Croatian Institute for Brain Research\\
  University of Zagreb, School of Medicine\\
  \texttt{goran.sedmak@hiim.hr} \\
  \And
  Sven Lon\v{c}ari\'{c} \\
  Faculty of Electrical Engineering and Computing\\
  University of Zagreb \\ 
  \texttt{sven.loncaric@fer.hr} \\
  \And
  Milo\v{s} Juda\v{s} \\
  Croatian Institute for Brain Research\\
  University of Zagreb, School of Medicine\\
  \texttt{milos.judas@hiim.hr} \\
}

\begin{document}
\maketitle

\begin{abstract}
In this paper, we present a novel method for analysis and segmentation of laminar structure of the cortex based on tissue characteristics whose change across the gray matter underlies distinctive  between cortical layers. We develop and analyze features of individual neurons to investigate changes in cytoarchitectonic differentiation and present a novel high-performance, automated framework for neuron-level histological image analysis. Local tissue and cell descriptors such as density, neuron size and other measures are used for development of more complex neuron features used in machine learning model trained on data manually labeled by three human experts.  Final neuron layer classifications were obtained by training a separate model for each expert and combining their probability outputs. Importances of developed neuron features on both global model level and individual prediction level are presented and discussed. 
\end{abstract}


\section{Introduction}
Interest in analysis of laminar structure is driven by the evidence of relationship between features of cytoarchitectonic structure and cortical functions. Today, it is thought that variations in neuronal distribution in the brain density determine function. These variations define several distinct layers of the cortex that are arranged parallel to the brain's folded surface. The structure of these layers, known as the laminar structure, also varies, as well as density and size of neurons found within the layers \cite{brodmann1909vergleichende}, \cite{von1925cytoarchitektonik}. Based on these variations, a series of spatially discrete areas of the brain are distinguished. The subtleties in this fine structure of the brain underlying it's function can be characterized in great detail by studying the organization of cells across the cortex \cite{kaas1993functional}. However, investigations in this field are mostly done manually, requiring significant amount of researchers' time, introducing observer-dependent bias and reducing reproducibility of the research. Computer-aided methods provide means for faster and more objective investigations of the cortical structure and enable researchers to gain better understanding of anatomical and functional organization of the brain, as well as studying many neurological and psychiatric diseases that cause subtle changes in the brain structure.

Since the first methods that introduced automation in analysis of cortical layers, the central idea was sampling of different tissue measures along transverse lines drawn either manually or semi-automatically \cite{SCHLEICHER1999165} across the cortex, perpendicular to the laminar structure and spanning the full width of the cortex \cite{hudspeth1976cytoarchitectonic}, \cite{ryzen1956microphotometric}, \cite{hopf1968registration}, \cite{zilles2002quantitative}. An important step in development of profile features was made in \cite{meyer2010number} by the introduction of automatic methods for estimation of cell counts, thus providing realistic information about neuron density. In \cite{wagstyl2018mapping}, authors combined profile information with machine learning methods to create laminar segmentation on the BigBrain \cite{amunts2013bigbrain} dataset. The first article that uses features and statistics of individual neurons appeared in 2017. In \cite{nosova2017automatic}, authors use automatic segmentation of cells in $7\mu$m thick Nissl-stained sections of the mouse brain and develop shape descriptor features for each cell. In the year 2018, a first approach that does not use profiles across the cortex was proposed \cite{li2019discrimination}. A combined approach of unsupervised and supervised machine learning was used on a dataset of 2-photon microscopic images of the rat cortex. Authors address the issue of human bias in manual segmentation of cortical layers and use an unsupervised clustering approach to identify and represent the laminar structure.

Over the years, the impact of human bias in brain parcellation has been increasingly recognized and many methods sought to overcome this issue by developing objective quantitative measures and usage of statistics to distinguish between different layers and brain areas. Recently available massive amounts of high-resolution and multimodal data are far beyond the capabilities of any kind of manual analysis. However, detailed anatomical and biologically meaningful information can be derived on a large-scale through the introduction of machine learning methods to the field that was until now dominated by the usage of exclusively image processing techniques. Derivation and study of higher-level tissue descriptors can reveal currently neglected structuring principles and provide deeper understanding of the laminar structure. This suggests that there is a need to move beyond limitations of manually created parcellation in conventional atlases towards data-driven analysis. However, the very recent methods are usually developed on rat or mouse cortex data, and the results for human cortex are lacking. 

\section{Materials and methods}
\begin{figure}[!ht]
	\centering
	\includegraphics[width=0.35\linewidth]{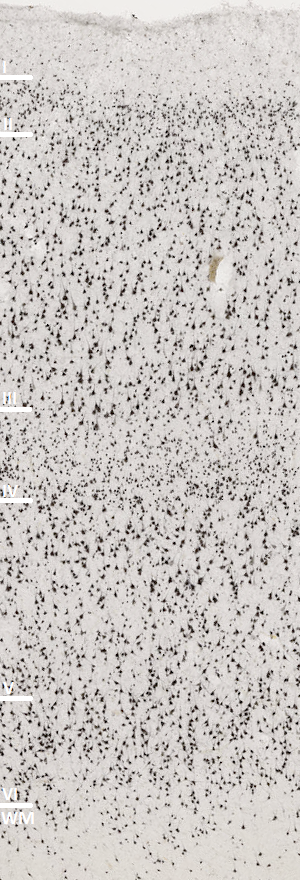}
	\includegraphics[width=0.35\linewidth]{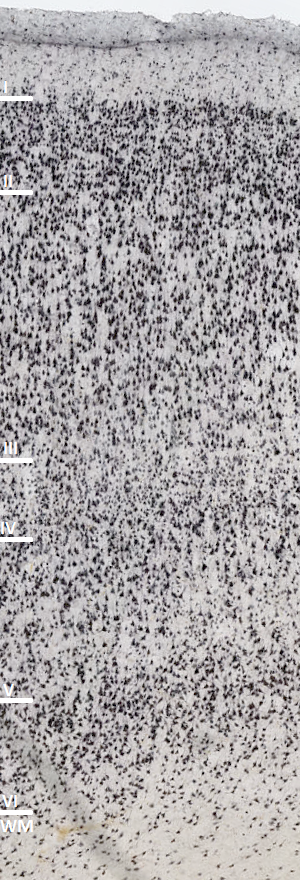}
	\captionof{figure}{Example of histological slices stained with NeuN immunohistochemistry method showing varying neuronal morphology and cellular distribution across the layers of the cortex.}
	\label{fig:samples}
\end{figure}

The data was obtained from the Zagreb Brain Collection \cite{judavs2011zagreb} and contained histological sections of the post-mortem adult human prefrontal cortex. All specimens were collected during regular autopsies at pathology departments of the University of Zagreb, School of Medicine, according to the protocol approved by the Institutional Review Board and with the next of kin consent. The tissue was stained using NeuN immunohistochemistry method. NeuN is a RNA-binding nuclear protein, derived from the RBFOX3 gene, which regulates alternative splicing in neurons and is specifically expressed in all neurons of used tissue specimens. Histological sections were digitized using Hamamatsu Nanozoomer 2.0 scanner (Hamamatsu Photonics, Japan), at 40x magnification, which corresponds to $0.226\mu$m/pixel resolution. In the experiments, $10\mu$m and $20\mu$m thick sections were used. Sections used in our experiments were obtained from medial portion of the ventral surface of orbitofrontal cortex, the \textit{gyrus rectus}, whose laminar structure is  composed of six cortical layers \cite{von1925cytoarchitektonik}, \cite{kaas1993functional}. Example histological sections in Fig. \ref{fig:samples} show varying neuronal morphology and cellular distribution across the layers of the cortex.

\begin{figure}[!ht]
	\centering
	\includegraphics[width=0.85\linewidth]{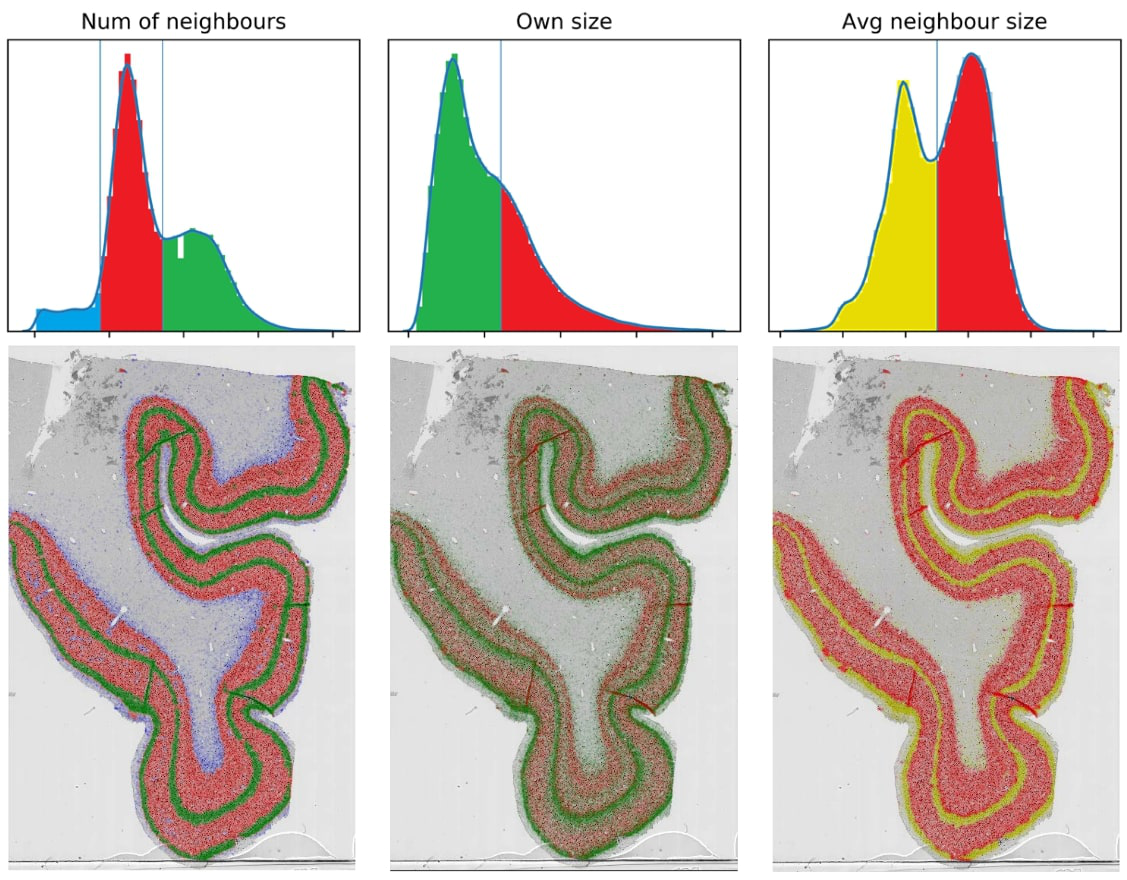}
	\captionof{figure}{Cell features exhibit multimodal distributions. Thresholds that separate the distributions were obtained using minimization of intraclass variance \cite{otsu1979threshold}. Left: Three types of neurons were distinguished by cell density in their surrounding area: very sparse (blue), sparse (red) and dense (green). Middle: Larger (red) and smaller (green) neurons. Right: Average size of neighboring neurons is a feature derived from the information about neuron's neighborhood and sizes of neurons found within and can also be used to facilite automatic layer segmentation.}
	\label{fig:classicalfeatures}
\end{figure}

According to classical works of cytoarchitectonics, neurons are usually distributed in six horizontally superimposed layers (layers I-VI) that are distinguished in most areas of the cortex \cite{brodmann1909vergleichende}, \cite{von1925cytoarchitektonik}, \cite{kaas1993functional}. Some neurons may be also found in the white matter (WM). This layered structure is caused by variations in cell density, size and shape of neurons, specific for each cortical layer. These, and potential variation in other features specific to either each individual neuron or its surroundings are the main focus in cortical layer segmentation. In manual delineation, density and size of neurons are the most important characterizations of the laminar structure. Based on anatomical descriptions and kernel density estimates, three populations of similar densities are assumed (layers II and IV as dense, layers III, V and VI as average, and Layer I and white matter as sparse), two populations of similar sizes (layers III, V and VI containing on average larger neurons, and layers I, II, IV and white matter containing on average smaller neurons). By plotting a histogram of neuron densities and sizes, one can observe that the features express a multimodal distribution, which can be separated using minimization of intraclass variance \cite{otsu1979threshold}. Fig. \ref{fig:classicalfeatures} shows visualization of separating the neuron populations across the histological section, revealing the laminar structure.

\begin{figure}[!ht]
	\centering
	\includegraphics[width=0.9\linewidth]{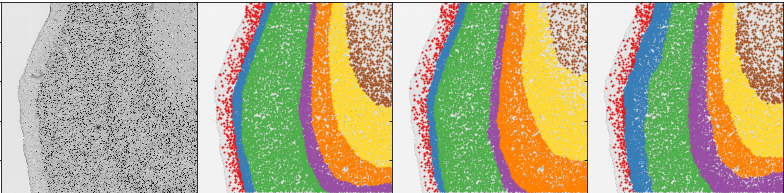}
	\captionof{figure}{Cortical layers manually delineated by the experts. Significant disagreement can be observed on the boundaries of the layers, as well as on positioning the boundary between the cortex and the white matter.}
	\label{fig:Expert123_plot}
\end{figure}

Automatic neuron detection and segmentation methods \cite{vstajduhar2018automatic}, \cite{stajduhar20183Dlocalization} were used to identify all neurons in the sections and obtain information about both locations and areas of neurons. Digitized histological sections were given to three human experts in histology and cytoarchitecture who manually delineated borders between the layers of the cortex. The apparent inconsistencies and mutual disagreement between the experts, as seen in Fig. \ref{fig:Expert123_plot}, shows the presence of experts' bias. The experts disagreed on the boundaries of all layers, with the exception of the very apparent layer I/layer II boundary.

In contrast to the classical pixel-based approach to cortical layer segmentation, we used  neuron-level tissue descriptors to characterize and examine the underlying tissue properties. We develop several feature classes that describe each neuron in the tissue and use a machine learning model to determine the layer of individual neurons. By classifying all neurons in the tissue, we obtain the parcellation of laminar structure. To the best knowledge of the authors, this is the first \textit{bottom-up} approach in analysis of brain cytoarchitecture  that builds from the cellular level and infers about larger structures based on features of
individual neurons.

Density-based clustering algorithms are often used to segment the areas of similar point densities \cite{rodriguez2014clustering}, \cite{ankerst1999optics}, \cite{campello2015hierarchical}. However, it seems that the neuron distribution in the cortex is such that their intrinsic structure may not be clustered by a single set of global density parameters. Nevertheless, clustering methods provided insight into some of its properties. Meaningful clusters were created when considering neurons within the radius between $100\mu$m and $300\mu$m, containing between $300$ and $800$ neurons. This leads to a conclusion that the changing nature of neuron distribution in the brain is best characterized when performing measurements in this range. Analysis of nearest neighbors is preferred over fixed radius approach, since a predefined radius may be interpreted differently, depending on the image resolution. Also, if a fixed number of neighbors for each neuron is used, efficient data structures like kd-trees \cite{bentley1975multidimensional}, \cite{maneewongvatana1999s} may be precomputed. Considering the large number of neurons found in a histological section, efficiency may be of critical importance.

\section{Results}

Neuron locations and segmentations were obtained using automated methods \cite{vstajduhar2018automatic}, \cite{stajduhar20183Dlocalization} which use grayscale-guided watershed on anisotropically diffused image to separate neurons, rather that distance maps obtained from gray-level threshold. The two steps provided a binary image of segmented, non-overlapping neuron areas. Combined with the original histological image, the two were used as an input to ImageJ particle analysis pipeline \cite{rueden2017imagej2} for computation of statistics and shape descriptors of segmentated neurons. Several basic features were computed for each neuron, including it's area, perimeter, as well as shape descriptors like circularity and roundness. These features form the basis for automated investigations in brain microanatomy. Some pixel intensity values such as mode, standard deviation, minimum and maximum gray level were not used in the analysis as these may be heavily influenced by uneven staining across the section and exhibit different values in different sections. 

\subsection{Neuron-level features}

\begin{figure}[!ht]
	\centering
		\includegraphics[width=\linewidth]{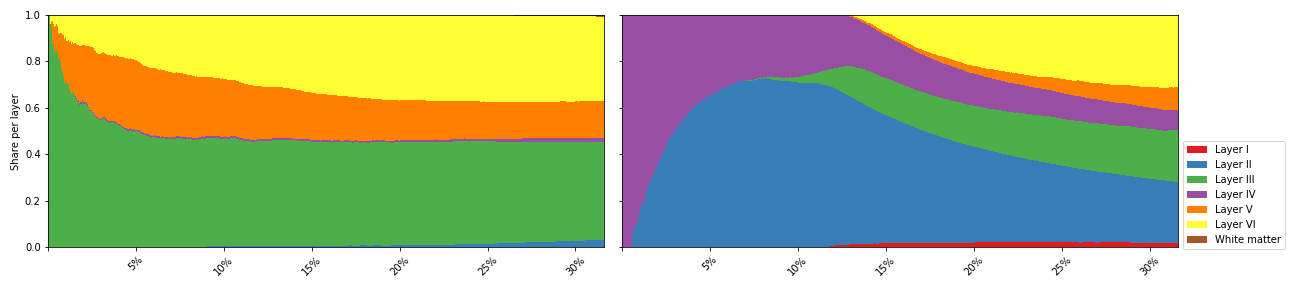}
	\caption{ Distribution of largest neurons per layer. Largest neurons were found in layer III of the cortex, and were followed by neurons of layer V and layer VI. Out of 50 largest neurons, $43 (86\%)$ were found in layer III,  $5 (10\%)$ in layer V and  $2 (4\%)$ in layer VI. Out of 500 largest neurons, $268 (54\%)$ were found in layer III,  $142 (28\%)$ in layer V,  $87 (17\%)$ in layer VI and only $3 (1\%)$ in layer IV, which follows neuroanatomical observations.}
	\label{fig:bodysizelayers}
\end{figure}

Distribution of values obtained in measuring various statistics of neurons across the cortex provides insight in different aspects of cytoarchitectural organization. This detailed, neuron-level approach allows for tissue inspection in accordance with known cytoarchitectonic principles like, for instance, distribution of the largest neurons. Those with largest \textit{area} were found in layer III of the cortex, and were followed by neurons of layer V and layer VI. Out of 50 largest neurons, $43 (86\%)$ were found in layer III,  $5 (10\%)$ in layer V and  $2 (4\%)$ in layer VI. Out of 500 largest neurons, $268 (54\%)$ were found in layer III,  $142 (28\%)$ in layer V,  $87 (17\%)$ in layer VI and only $3 (1\%)$ in layer IV. This comparison confirms that the automated approach yields meaningful results and follows neuroanatomical observations. Visualization of the ratio of the distribution of largest $500$ neurons among the layers is shown in Fig. \ref{fig:bodysizelayers}. Neuron \textit{circularity} and \textit{roundness} were found the lowest in layer VI which is known to consist of multipolar neurons with dendrites reaching in different directions. Variations in grayscale intensity were expressed differentially in the cortical layers. Neurons with highest mean grayscale values were mostly found in layer I, showing low NeuN dye intake. Neurons with lowest \textit{median} were predominantly found in layer VI, in layer IV and in the middle of layer III, sometimes referred to as layer IIIb. No conclusion was made or the reason found for neurons of layer VI having such large NeuN uptake properties that resulted in lower individual grayscale intensities.  

For the distances to a neurons' $k$-th nearest neighbor, mean, skewness, kurtosis, entropy and other statistics were computed for $k \in [50,100,250,500,1000]$. Measures regarding neuron shape such as area, circularity or perimeter provided more discriminative power, which is not unexpected, since the findings in neuroanatomical research relies to large extent on shape and size of neurons. A convex hull of neuron's $k$-neighbors gives information about area around a neuron and a number of its neighbors and is described using hull area, perimeter, average nearest distance for neurons found in the hull and standard deviation of nearest distances. 
Dispersion of neurons is quantified using \textit{nearest neighbor index} (NNI), a measure which describes whether points follow usually subjective patterns of regular, clustered or random distribution. 
NNI measures the distance between each point and its nearest neighbor's location. All the nearest neighbor distances are averaged, and if the average distance is less than the average for a random distribution, the distribution of the features being analyzed is considered clustered. If the average distance is greater than a random distribution, the features are considered regularly dispersed. The index is expressed as the ratio of the mean observed distance divided by the expected distance, which is based on a random distribution with the same number of points covering the same total area,
\begin{equation}
NNI_i = \displaystyle\frac{\frac{1}{n} \sum_{j=1}^n d(i,j)}{0.5 \sqrt{HullArea(i)/n}}.
\end{equation}
Neurons in all layers with the exception of layer I and white matter tend more towards uniformly dispersed distribution, especially neurons of layer IV which tend more towards random distribution. 

Depending on it's position in the cortex, a neuron may be placed more towards middle or more towards the edge of its layer. Computation of properties of its neighborhood may be confounded by reaching into adjacent layer and using neurons with different propertied for computation of statistics. To identify this case, measurements may be taken only from neurons found within the range of angle, or \textit{slices}. Features measured in several directions can identify border neurons and changes of neuronal properties in different directions. Slices may be regarded as measurement units reaching from a single neuron, each unit representing a population of neighboring neurons found in a given direction from the central neuron. The relationship of different populations within an area has been extensively studied in the frame of biological diversity of species, landscapes and other \cite{magurran2013measuring}, \cite{nagendra2002opposite}. Considering the neurons in a slice as members of a single \textit{species}, and the $k$ neighbors of a neuron as the population of all species in their habitat,  biodiversity measures evaluate the relationship between the species. 
The two most often used such measures are Shannon index \cite{shannon1948mathematical} and Simpson index \cite{simpson1949measurement}. The Shannon index gives a quantitative measure of the uncertainty in predicting the species of an individual chosen randomly from the population. The Simpson index measures the probability that the two individuals randomly chosen (with replacement) from the total population will be of the same species.
\begin{equation}
Shannon = - \sum_{i=1}^{R} p_i \ln p_i = 
\ln \left( \frac{1}{\prod_{i=1}^{R} p_i^{p_i}} \right),
\quad
Simpson = \sum_{i=1}^{R} p_i^2,
\label{eq:shannonsimpson}
\end{equation}
where $R$ is the number of different species, or slices, and $p_i$ is the proportion of species of the $i$th type in the population, or proportion of neurons in $i$th slice to the number of the neurons in $k$-neighborhood. If all slices have equal number of neurons, $p_i$ values equal $1/R$, and the Shannon index takes the maximum value of $\ln R$. If the numbers are unequal, the weighted geometric mean of the $p_i$ values is larger, which results in the index having smaller values. The index equals zero if the neurons from only one slice are present, since there is no uncertainty in predicting the slice they are in. The index gives information about the relation of number of types and the presence of the dominant type. Mean proportional abundance of the slices increases with decreasing number of slices and the increasing abundance of the slice with largest number of neurons, the index obtains small values in regions of high diversity like neurons on borders between the layers, thin layers, and especially layer I neurons. The index is large in homogeneous areas like the middle of layer III, where slices reaching from a neuron remain in the area of the layer. 

\begin{figure}[!ht]
	\centering
	\includegraphics[width=0.9\linewidth]{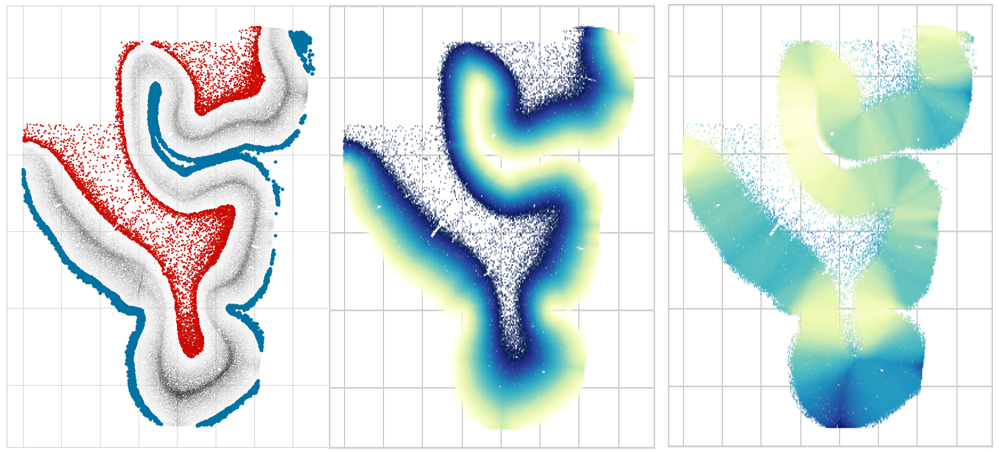}
	\caption{Features based on local density and convex hull radius are used to obtain tissue features without drawing profiles and sampling perpendicular to the cortex. Left: sparse areas are separated into layer I and white matter. Middle: cortical depth. Right: cortical thickness.   }
	\label{fig:depththickness}
\end{figure}

Using local neuron density, layer I and white matter can be distinguished by having small neuron density, thus identifying \textit{sparse regions} of the section, or \textit{dense regions} containing layers II and IV, as shown in Fig. \ref{fig:classicalfeatures}. The sparse region may further be split using the hull area feature - neurons in white matter will have large hull area, in contrast to the neurons of layer I, whose hull is bound between the border of the tissue and the dense layer II. By computing distances to layer I and white matter, cortical thickness and depth of each neuron are derived, as shown in Fig. \ref{fig:depththickness}.

\subsection{Machine learning pipeline}
Developed neuron feature sets provide quantitative descriptors of cortical organization. No single feature was found that would be able to provide clear segmentation of cortical layers and although some features were more expressed in certain layers than the others, it is not straightforward what exactly is changing between the layers as well as the impact and interconnection of different features. This lead to an assumption that there is information contained in the developed features that can be analyzed, combined and used to produce precise classification of neurons with regard to their location within the cortical layers. A classical supervised machine learning approach is used on a dataset of manually segmented layers. We develop a model that predicts the neurons' layers and produces the segmentation of cortical layers on the whole histological section. Feature attributions for the model are investigated to identify important tissue descriptors.

\begin{figure}[!ht]
	\centering
	\includegraphics[width=\linewidth]{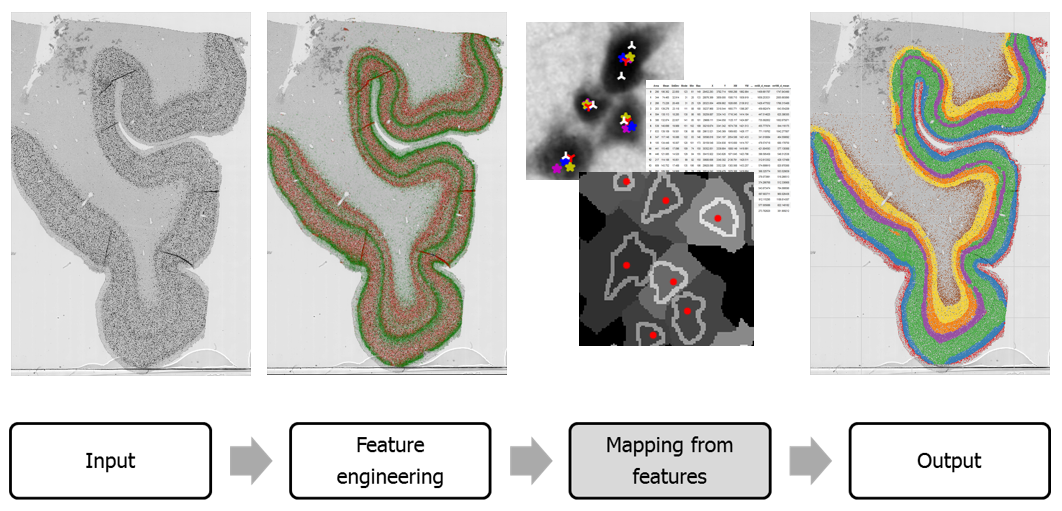}
	\captionof{figure}{Cortical layer segmentation through a classical machine learning pipeline. The hand-crafted neuron features based on automatic segmentation were used as an input to a machine learning model which learned to map the neurons' features to cortical layers. Machine learning model learned the variations in neuronal features and create consistent predictions of neurons' layers across the whole histological section. Black rectangle frames the data that was manually labeled by the experts and used for training. }
	\label{fig:pipeline}
\end{figure}

Boosted decision trees, a state-of-the-art method for prediction problems with input data such as the computed neuron features \cite{bramer2007principles}, \cite{liu2017towards} were chosen for prediction and interpretation of cortical lamination for its several advantages. Decision trees mirror human decision making more closely than other approaches \cite{james2013introduction}, which is especially useful when modeling human activities, such as manual delineation of cortical layers, a decision-making process based on combination of several information about neurons' characteristics. We used CatBoost \cite{dorogush2018catboost}, a method based on gradient boosting over decision trees. The best generalizations were obtained by combining the manual labels of all three raters. The model was trained for each rater using \texttt{softmax} objective, output probabilities were summed, and the final prediction was made using a maximum over all classes for each instance. Results of this approach are shown in the most right of Fig. \ref{fig:pipeline}. Classes of neurons are predicted, and neurons are grouped in a way that follows laminar pattern of the cortex.  



Experiments with different sets of features have shown that although some combinations of features do achieve high accuracy on training data, that does not guarantee that the model will perform well on the whole histological section. The introduction of features based on distance to sparse or dense regions has significantly improved model's ability to separate regions of the sections into parcels following the laminar layout of the cortex. 


\subsection{Comparison with human experts}
Without the existence of a single ground truth for reference, the measurement of model's performance is considered in the context of inter-rater variability. Training data was split in $75\%$ training and $25\%$ test subsets, and predictions of the model were compared with the experts' manual labels. Comparing neuron layer prediction, average agreement between two experts was $0.7549 \pm 0.0485$ for $10\mu$m and $0.8090 \pm 0.0487$ for $20\mu$m histological section. Average accuracy of the model when compared to the three experts was $0.8724\pm 0.0419$ and $0.8967 \pm 0.0465$.

\subsection{Analysis of individual feature attribution}
\begin{figure}[!ht]
	\centering
	\includegraphics[width=0.6\linewidth]{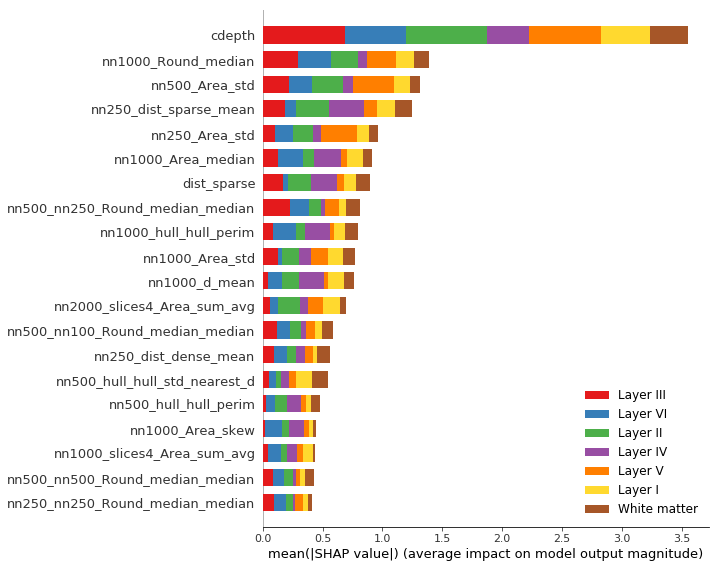}
	\caption{ Importance of neuron features at the model level using the SHAP feature importance analysis. }
	\label{fig:shap_importances}
\end{figure}

For deeper understanding of both the model and features being used in the pipeline, investigation of the impact of features on prediction of neuron's class was performed on both global (model) and instance (individual neuron) level. A recent approach for measuring feature attribution, the SHAP measure \cite{NIPS2017_7062}, was used for estimation of neuron features that facilitate neuron classification within the layers. Importance of the features used for training the ensemble model are shown in Fig. \ref{fig:shap_importances}.

\begin{figure}[!ht]
	\centering
	\includegraphics[width=\linewidth]{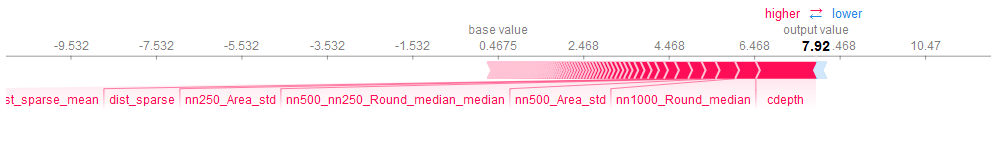}
	\includegraphics[width=\linewidth]{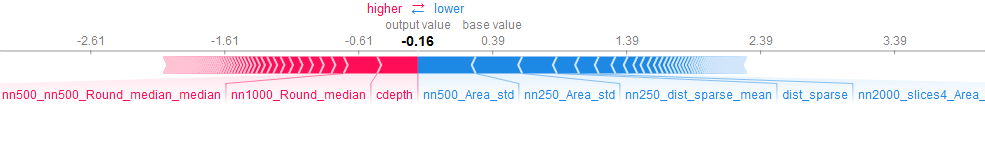}
	\caption{ Contribution of different features for making a prediction of a neuron's layer. Top: neuron features of a single neuron of layer III that contributed to increase from the base SHAP value and making the prediction. Bottom: the importance of features that decreased the output value for prediction of the same neuron as a layer II neuron.}
	\label{fig:shap_individual}
\end{figure}

An important aspect of this approach is the ability of identifying features that contribute to making a prediction on a single instance of the data, for each individual neuron. Fig. \ref{fig:shap_individual} shows which neuron features for a neuron of layer VI contributed to increasing the base SHAP value and making the prediction. The image also shows the impact of features that decreased the output value for prediction of the same neuron as a white matter neuron.

It was shown in several occasions that this findings really follow neuroanatomical findings
pretačemo ih iz ljudskih opažanja u računalna
Površina neurona nije bitna, nego što se događa u okolini
na primjer, istu površinu mogu imati piramidni i interneuron, no važan je oblik
Roundness je veći u 5/6 sloju
()izbaciti sliku 7) i staviti novu u pipeline
prednost ovog pristupa je što ne daje ravne nego vrludave granice među slojevima, koje su anatomski smislenije

 Cortical location features had largest impact on the model's output. This is because they integrate low level features with anatomical observations, and have a wide reach across the histological section. Features based on oriented measurements that measure change of cytoarchitectonic properties in different directions also had considerable feature importance, being able to identify neurons on the border of cortical layers. It was shown that building on lower-level features yields features that have greater discriminative power and thus greater importance. This is because of their capacity to overcome local variations in neuron features and taking, for instance, mean of those features. In contrast, using features of a neuron such as area, there is no increase in accuracy of prediction, which is reflected in this features having minimal importance. This is probably why different methods for local pattern analysis and classical image feature extraction methods are not very successful in cortical layer segmentation. Range of variability radius was established, giving an estimate of the size of the neuron's neighborhood in which measurements should be made in order for it to be large enough to overcome local variations in neuron distribution and recognize its location within the cortical structure on one hand, and on the other narrow enough so that measurements are not confounded by reaching too far into adjacent layers.

During experiments with different models and combinations of datasets, it was noticed that parts of layer III and layer VI are sometimes divided into sub-layers that follow the direction of laminar structure, although being very short and not extending through a significant portion of the slice. Further investigations using the developed methodology may provide more detailed insight into sub-layering of cortical structure.

\section{Conclusion}
This paper presents a novel framework for analysis of laminar structure of the cortex. From the location and basic measures of neurons, more complex features were developed and used in machine learning pipeline. Tree ensembles, as today's one of the most powerful and interpretable models were used on data manually labeled by three human experts in order to account for intra-rater variations. Classification results were obtained by training three CatBoost models separately on dataset labeled  by each experts and creating an ensemble, combining their probability outputs in final neuron class prediction. Importance of developed features on both model-level and instance-level was performed using SHAP analysis.

\subsection*{Acknowledgments.}
This publication was supported by the European Union through the European Regional Development Fund, Operational Programme Competitiveness and Cohesion, grant agreement No. KK.01.1.1.01.0007, CoRE - Neuro; and the Canada First Research Excellence Fund, awarded to McGill University for the Healthy Brains for Healthy Lives initiative.

Authors extend their gratitude to Dora Sedmak from Croatian Institute for Brain Research (CIBR), School of Medicine, University of Zagreb, and Jennifer Novek from Montreal Neurological Institute (MNI), McGill University, for their effort in neuron labeling and helpful discussions. Special thanks to Claude Lepage from the MNI, McGill University, for reading the paper thoroughly and providing constructive feedback.

\bibliographystyle{unsrt}
\bibliography{bibliography}

\end{document}